\documentclass[runningheads]{llncs}
\usepackage[T1]{fontenc}
\usepackage{graphicx,verbatim}
\usepackage{cite}
\usepackage[colorlinks=true,citecolor=blue,linkcolor=blue,urlcolor=blue]{hyperref}
\usepackage{amssymb} 
\usepackage{amsmath}
\usepackage{amsfonts}
\usepackage{booktabs}
\usepackage{multirow} 
\usepackage[most]{tcolorbox}
\usepackage{pifont}
\usepackage{adjustbox}
\begin{document}
\title{Synergistic Perception-Reasoning Governance: Grounding Medical MLLMs with Verifiable Anatomical Evidence}
\titlerunning{Grounding Medical MLLMs with Verifiable Anatomical Evidence}
%
\author{Rui Hao\inst{1}\and
Qiankun Li\inst{2,3}\textsuperscript{\(\dagger\)}\and
Junyuan Mao\inst{4}\and Linghao Meng\inst{4}\and Dirui Xie\inst{1}\and Dayu Tan\inst{5}\and Zhigang Zeng\inst{1}\textsuperscript{\(\dagger\)}}
%

\authorrunning{Rui Hao et al.}
%
\institute{
Huazhong University of Science and Technology, Wuhan, China
\and
Imperial Global Singapore, Imperial College London, London, United Kingdom
\and
Nanyang Technological University, Singapore, Singapore
\and
National University of Singapore, Singapore, Singapore
\and
Anhui University, Hefei, China\\
\email{
ruihao@hust.edu.cn;
q.li2@imperial.ac.uk;
zgzeng@hust.edu.cn
}
\textsuperscript{\(\dagger\)}Corresponding author.
}
\maketitle              
\begin{abstract}
Multimodal large language models (MLLMs) show strong promise for clinical VQA and radiology report generation, yet inference-time hallucinations still undermine trustworthy use: models can produce fluent conclusions that conflict with imaging evidence. Existing mitigation strategies typically rely on additional training, external retrieval/knowledge bases, or multi-stage post-hoc verification, which increases cost and pipeline complexity and often generalizes poorly across models and tasks.
To address this, we propose a holistic, training-free evidence-injection framework that systematically mitigates hallucinations through dual-side evidence injection. By leveraging ROI priors acquired using MedSAM in our implementation, we recalibrate the visual perception trajectory via ROI-guided activation modulation while anchoring the textual reasoning trajectory by mapping anatomical coordinates into discrete semantic tokens as verifiable external memory. Then we introduce a task-aware dynamic router to select modality-specific interventions based on task semantics, balancing perceptual grounding and linguistic fluency. We conduct systematic evaluations on 2 tasks and 5 datasets using \texttt{LLaVA-1.5-7B}, \texttt{LLaVA-Med-1.5-7B}, \texttt{Qwen3-VL-8B/32B}, and \texttt{InternVL-3.5-8B/38B}. Controlled ablations and visualizations further validate the framework, which consistently outperforms baselines across medical benchmarks, improving close-ended accuracy by up to $\sim\mathbf{6}\%\uparrow$ and reducing open-ended hallucinations by $\sim\mathbf{35}\%\downarrow$. The code has been made available on GitHub: \href{https://github.com/Henry991115/SPRG}{\textcolor{blue}{https://github.com/Henry991115/SPRG}}.


\end{abstract}
\section{Introduction}
MLLMs combine visual perception and language generation with emerging clinical reasoning, enabling medical VQA and radiology report generation \cite{chen2024chexagent,moor2023medflamingo,thawakar2024xraygpt,wu2025radfm,lee2024llmcxr}. However, hallucinations remain a key barrier to trustworthy deployment \cite{jing2025tutorialhallucinations}, as models may generate fluent outputs that are unsupported by or even contradict imaging evidence \cite{lin2024trustworthyreview,tu2025attentionreallocation}. Recent benchmarks categorize medical hallucinations into visual misinterpretation, knowledge deficiency, and context misalignment \cite{chang2025medheval}; in high-stakes settings, such evidence-inconsistent outputs can lead to incorrect lesion judgments and unsafe clinical attributions \cite{zhu2025trust,asgari2025clinicalsafety}.

To mitigate these risks, many training-free, inference-time methods have emerged. VCD \cite{leng2024vcd}, DoLa \cite{chuang2023dola}, and OPERA \cite{huang2024opera}adjust decoding by contrasting perturbed inputs, contrasting layer logits, or penalizing over-trusted tokens. AVISC \cite{woo2024avisc}, DAMRO \cite{gong2024damro}, and PAI \cite{liu2025pai}calibrate attention or visual tokens by correcting outliers, filtering vision-side noise, and reducing text-dominant bias. M3ID \cite{favero2024m3id} strengthens image grounding through visually dominant constraints. However, most existing mitigations rely on internal statistical calibration without explicit anatomical grounding, which limits generalization across clinical tasks with different evidence requirements.

In this work, we propose a holistic, training-free inference-time evidence-injection framework to mitigate visual misinterpretation hallucinations in medical MLLMs. Using ROI priors as verifiable anatomical evidence, we improve visual grounding and clinical consistency through the perception-reasoning intervention mechanism, including text-side evidence injection and vision-side activation recalibration over visual tokens, with a joint strategy for complementary gains. To handle heterogeneous clinical tasks, we further introduce a lightweight task-aware dynamic router that dynamically selects modality-specific interventions based on task semantics. Extensive experiments across diverse medical benchmarks, supported by controlled ablations and visualizations, show superior robustness over existing baselines, with up to $\sim \mathbf{6}\mathbf{\%} \mathbf{\uparrow}$ improvement in close-ended reasoning accuracy and up to $\sim \mathbf{35}\mathbf{\%} \mathbf{\downarrow}$ reduction in open-ended hallucinations. Our key contributions can be summarized as follows:
\begin{itemize}
\item[\ding{182}] We introduce a novel training-free framework that mitigates hallucinations by injecting dual-side verifiable anatomical evidence, bridging the gap between raw visual perception and clinical reasoning.
\item[\ding{183}] We develop a perception-reasoning intervention mechanism that recalibrates the visual perception trajectory via ROI-guided visual-side activation modulation, while mapping anatomical coordinates into discrete semantic tokens as verifiable external memory to anchor the textual reasoning trajectory.
\item[\ding{184}] We propose a task-aware dynamic router that adjusts modality-specific intervention strategies according to task semantics, thereby achieving a better balance between perceptual grounding and linguistic fluency.
\item[\ding{185}] We establish a rigorous cross-model, cross-task evaluation framework involving 2 tasks, 5 datasets and 6 MLLMs, providing a highly comprehensive study to date on inference-time medical hallucination mitigation.
\end{itemize}

\section{Methods}
\label{sec:methods}

\begin{figure}[t]
    \centering
    \includegraphics[width=1.0\linewidth]{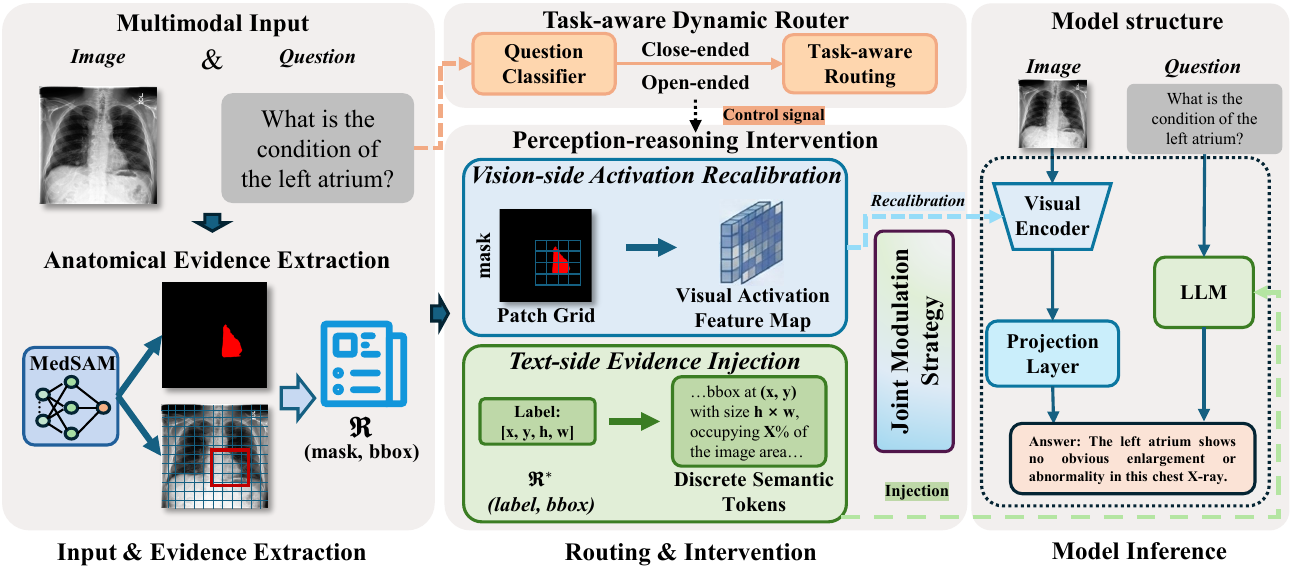}
    \caption{The overall pipeline consists of all input, ROI prior extraction, intervention mechanism, task-aware routing, and model inference for the output generation.}
    \label{fig:overview}
\end{figure}

Given an image $I$, a textual question $q$, and a pretrained MLLM parametrized by $\theta$, the standard autoregressive generation is formulated as:
\begin{equation}
 p_{\theta}(y \mid I, q) = \prod_{t=1}^{T} p_{\theta}\!\left(y_t \mid y_{<t}, \mathrm{Enc}(I), q\right),
\end{equation}
where $\mathrm{Enc}(I) \in \mathcal{R}^{N \times d}$ denotes the sequence of $N$ visual patch tokens. To mitigate visual misinterpretation, we introduce an intervention function that conditionally modulates the visual representation $\tilde{V}$ and textual prompt $\tilde{q}$ based on verifiable anatomical priors $\mathcal{R}$, yielding $p_{\theta,\phi}(y \mid I, q, R) = \prod_{t=1}^{T} p_{\theta}(y_t \mid y_{<t}, \tilde{V}, \tilde{q})$.

As shown in Fig.~\ref{fig:overview}, the pipeline comprises the anatomical evidence extraction, perception-reasoning intervention mechanism, and task-aware dynamic router.

\subsection{Anatomical Evidence Extraction}
We utilize MedSAM prompted by a candidate bounding box set $\mathcal{B}=\{b_k\}_{k=1}^{K}$ to extract verifiable anatomical evidence. For each valid box $b_k$, we obtain its corresponding segmentation mask $m_k$. After deduplication and invalid region filtering, we define the deterministic ROI prior set as $\mathcal{R}=\{(m_k, b_k, l_k)\}_{k=1}^{K'}$, where $l_k$ is the anatomical semantic label derived from the clinical context.

\subsection{Perception-reasoning Intervention Mechanism}

\paragraph{\textbf{Text-side Evidence Injection.}}
We convert the ROI prior set $\mathcal{R}$ into a collection of verifiable region descriptors and inject them into the textual prompt in a structured form. 
Given $\mathcal{R}=\{(m_k,\hat b_k)\}_{k=1}^{K'}$, we represent each region as a detection pair $(l_k,b_k)$, where $l_k$ is the anatomical label and $b_k=[x,y,w,h]$ specifies its location and scale. We then sequentially construct the structured region set, the serialized evidence block, and the injected prompt as
\begin{equation}
\tilde{q} = \left[ \mathcal{S}(\mathcal{R}) \parallel q \parallel \eta_{\mathrm{task}} \right], \quad \tilde{V} = V,
\end{equation}
where $\mathcal{S}(\cdot)$ is a deterministic serialization operator. The operator $\parallel$ denotes token-level concatenation, and $\eta_{\mathrm{task}}=\texttt{``Answer Yes/No''}$ is optionally appended for binary QA to reduce format drift. The final prediction is generated by conditioning on $(I,\tilde q)$. 

\paragraph{\textbf{Vision-side Activation Recalibration.}}
Let $\mathcal{P}_i \subset \mathbb{Z}^2$ denote the discrete spatial domain of the $i$-th visual patch token, where $i \in \{1, \dots, N\}$. We compute the spatial coverage density $\gamma_i$ by aggregating the global binary ROI mask $\mathbf{B}(u,v) = \bigvee_{k} m_k(u,v)$ over the patch, normalized by its cardinality $|\mathcal{P}_i|$:
\begin{equation}
    \gamma_i = \frac{1}{|\mathcal{P}_i|} \sum_{(u,v) \in \mathcal{P}_i} \mathbf{B}(u,v).
\end{equation}

To suppress out-of-distribution (OOD) structural collapse while extinguishing background noise, we formulate an activation soft-mask $\mathbf{m} \in \mathbb{R}^N$ governed by the threshold indicator $\mathbb{I}(\cdot)$:
\begin{equation}
    m_i = \alpha \cdot \mathbb{I}(\gamma_i \ge \rho) + \beta \cdot \left(1 - \mathbb{I}(\gamma_i \ge \rho)\right),
\end{equation}
where $\rho$ is the density threshold, $\alpha$ serves as a gain factor about the ROI-aligned features, and $\beta \in (0, 1)$ acts as a severe $L_2$-norm attenuation factor. Implemented via a forward hook prior to the cross-modal projection layer, the recalibration is executed via broadcasting scalar multiplication: 
\begin{equation}
    \tilde{V} = \mathbf{Diag}(m) V, \quad \tilde{q} = q.
\end{equation}

By shrinking the feature norm of non-ROI patches by a factor of $\beta$, their subsequent dot-product attention scores approach zero exponentially, effectively excising spurious visual pathways.

\paragraph{\textbf{Joint Modulation Strategy.}}
To achieve complementary regularization, the joint mapping simultaneously enforces $L_2$-norm visual filtering and discrete semantic anchoring:
\begin{equation}
\Phi_{\mathrm{Joint}}: \quad \tilde{V} = \mathbf{Diag}(m) V, \quad \tilde{q} = \left[ \mathcal{S}(\mathcal{R}) \parallel q \parallel \eta_{\mathrm{task}} \right].
\end{equation}

\subsection{Task-aware Dynamic Router}
Since distinct clinical question categories exhibit varying sensitivities to visual vs. textual constraints, statically applying a single strategy yields sub-optimal robustness. We therefore propose a zero-overhead dynamic router based on zero-shot LLM question classification at inference.

Given question $q$, a lightweight classifier yields its category $\hat{c} = h(q) \in \{\mathcal{C}_{A}, \mathcal{C}_{M}, \mathcal{C}_{S}, \mathcal{C}_{R}\}$, corresponding to Anatomy, Measurement, Symptom, and Radiology features, respectively. We define the routing strategy dynamically based on the query category: $s(\hat{c}) =$ \textbf{Vision-side} if $\hat{c} \in \{\mathcal{C}_{A}\}$, \textbf{Text-side} if $\hat{c} \in \{\mathcal{C}_{M}, \mathcal{C}_{S}\}$, and \textbf{Joint} if $\hat{c} \in \{\mathcal{C}_{R}\}$.

\textbf{Rationale:} Anatomy ($\mathcal{C}_{A}$) relies on spatial localization, naturally benefiting from vision-side L2-norm recalibration. Measurement ($\mathcal{C}_{M}$) and Symptoms ($\mathcal{C}_{S}$) demand precise semantic anchoring, making structured textual prompts superior. Radiology features ($\mathcal{C}_{R}$) typically require complex cross-modal reasoning, necessitating joint constraints on both visual activations and textual generation. 

\section{Experiments and Results}
\label{sec:Experiments and Results}
\subsection{Datasets and Evaluation Metric}
We evaluate hallucinations following the benchmark by Chang et al.~\cite{chang2025medheval}. For close-ended tasks, we report overall and per-type accuracy (anatomy, measurement, symptom, radiology) on MM-VisHal (comprising SLAKE~\cite{liu2021slake} and VQA-RAD~\cite{lau2018vqarad}) and CXR-VisHal (featuring IU-Xray~\cite{demner2016iu} and MIMIC-CXR~\cite{johnson2019mimiccxr}). For open-ended evaluation, we assess hallucinations and text quality via CHAIR, CheXbert, RadGraph, RaTEScore, and Recall using 490 image-report pairs sampled from the MIMIC-CXR test set.

\begin{table}[t]
\centering
\caption{Close-ended evaluation results on MM-VisHal and CXR-VisHal datasets. A, M, S, and R denote Acc-A, Acc-M, Acc-S, and Acc-R, respectively. (\textbf{Note:} Due to space constraints, MM-VisHal includes SLAKE and VQA-RAD, while CXR-VisHal comprises IU-Xray and MIMIC-CXR.)}
\label{tab:close_vqa_main}
\begin{tabular}{c|c|ccccc|ccccc}
\hline
\multirow{2}{*}{\textbf{MLLM}} & \multirow{2}{*}{\textbf{Method}} &
\multicolumn{5}{c|}{\textbf{MM-VisHal}} &
\multicolumn{5}{c}{\textbf{CXR-VisHal}} \\
\cline{3-12}
&  & A & M & S & R & Acc & A & M & S & R & Acc \\
\hline
\multirow{4}{*}{\shortstack{LLaVA-1.5-\\7B}} & Original & 58.5 & 42.3 & 55.7 & 50.8 & 52.3 & 73.2 & 43.0 & 52.7 & 58.2 & 55.7 \\
  & Text-side & 65.6 & \textbf{44.1} & \textbf{57.3} & 51.3 & \textbf{55.0} & 73.6 & \textbf{45.2} & \textbf{55.9} & 56.9 & \textbf{57.9} \\
  & Vision-side & \textbf{69.8} & 42.9 & 54.7 & 52.7 & \textbf{55.0} & \textbf{74.7} & 38.6 & 54.3 & 58.3 & 56.5 \\
  & Joint & 61.7 & 42.8 & 56.2 & \textbf{52.8} & 53.6 & 72.4 & 41.7 & 54.9 & \textbf{59.1} & 56.9 \\
\hline
\multirow{4}{*}{\shortstack{LLaVA-Med-1.5-\\7B}} & Original & 65.9 & 38.2 & 51.1 & 53.2 & 51.7 & 85.3 & 46.3 & 70.7 & 85.3 & 71.9 \\
  & Text-side & 65.9 & \textbf{40.1} & \textbf{57.3} & 52.8 & \textbf{54.4} & 86.2 & 48.9 & \textbf{73.3} & 86.6 & \textbf{74.1} \\
  & Vision-side & \textbf{66.9} & 39.1 & 53.6 & 51.8 & 52.8 & \textbf{86.5} & 48.1 & 71.3 & 85.6 & 72.7 \\
  & Joint & 65.3 & 37.9 & 55.0 & \textbf{54.6} & 53.2 & 85.2 & \textbf{49.2} & 71.5 & \textbf{87.6} & 72.9 \\
\hline
\multirow{4}{*}{\shortstack{Qwen3-VL-\\8B}} & Original & 73.1 & 61.7 & 66.0 & 72.0 & 67.6 & 77.0 & 46.8 & 78.5 & 80.7 & 75.0 \\
  & Text-side & 76.9 & \textbf{65.1} & 70.2 & \textbf{72.7} & \textbf{70.9} & 77.8 & \textbf{50.4} & \textbf{79.7} & 81.7 & \textbf{76.3} \\
  & Vision-side & \textbf{77.9} & 63.1 & 65.7 & 72.2 & 69.0 & \textbf{78.3} & 48.0 & 78.3 & 82.3 & 75.4 \\
  & Joint & 72.9 & 63.0 & \textbf{70.5} & \textbf{72.7} & 69.6 & 78.1 & 50.0 & 78.0 & \textbf{83.6} & 75.5 \\
\hline
\multirow{4}{*}{\shortstack{InternVL3.5-\\8B}} & Original & 76.6 & 65.0 & 70.4 & 84.4 & 72.9 & 88.1 & 73.5 & 85.1 & 93.7 & 85.1 \\
  & Text-side & \textbf{80.2} & \textbf{66.7} & \textbf{73.6} & 84.4 & \textbf{75.3} & \textbf{88.5} & 74.5 & 85.4 & 92.9 & 85.4 \\
  & Vision-side & 78.1 & 65.2 & 70.5 & \textbf{84.8} & 73.4 & 88.4 & 73.5 & \textbf{85.7} & 94.1 & \textbf{85.6} \\
  & Joint & 78.6 & 66.0 & 73.0 & 84.1 & 74.5 & 87.8 & \textbf{74.8} & 85.3 & \textbf{94.5} & 85.4 \\
\hline
\multirow{4}{*}{\shortstack{Qwen3-VL-\\32B}} & Original & 78.6 & 68.3 & 66.6 & 74.7 & 71.1 & 79.0 & 60.7 & 81.9 & 79.2 & 78.8 \\
  & Text-side & 79.5 & \textbf{69.9} & \textbf{70.6} & \textbf{76.7} & \textbf{73.5} & 78.8 & \textbf{62.5} & \textbf{83.1} & 76.7 & \textbf{79.5} \\
  & Vision-side & \textbf{80.3} & 68.0 & 69.5 & 74.8 & 72.5 & \textbf{79.7} & 62.2 & 81.8 & 79.8 & 79.1 \\
  & Joint & 77.9 & 67.4 & 70.5 & 75.9 & 72.4 & 79.3 & 61.9 & 82.3 & \textbf{81.5} & \textbf{79.5} \\
\hline
\multirow{4}{*}{\shortstack{InternVL3.5-\\38B}} & Original & 79.1 & 69.4 & 70.8 & 81.2 & 74.1 & 80.4 & 71.4 & 86.3 & 91.7 & 84.2 \\
  & Text-side & 79.8 & 70.2 & \textbf{74.8} & 81.0 & \textbf{75.9} & 80.9 & \textbf{75.6} & \textbf{87.0} & 91.6 & \textbf{85.1} \\
  & Vision-side & \textbf{81.0} & \textbf{70.5} & 71.9 & 81.4 & 75.3 & \textbf{81.2} & 72.8 & 86.4 & 91.7 & 84.5 \\
  & Joint & 80.7 & 67.3 & \textbf{74.8} & \textbf{82.6} & 75.6 & 81.1 & 72.5 & 86.5 & \textbf{93.1} & 84.7 \\
\hline
\end{tabular}
\end{table}

\subsection{Results}
On both close-ended and open-ended evaluation tasks, we systematically evaluate LLaVA-1.5-7B \cite{liu2024llava15}, LLaVA-Med-1.5-7B \cite{li2023llavamed}, Qwen3-VL-8B/32B \cite{bai2025qwen3vl}, and InternVL-3.5-8B/38B \cite{wang2025internvl35}; the following sections present these results in turn.

\paragraph{\textbf{Close-ended Evaluation.}}
Table~\ref{tab:close_vqa_main} shows that our training-free evidence injection consistently improves accuracy across all MLLMs. Notably, text-side injection dominates, achieving the highest overall Acc in $9$ of $12$ settings (with $2$ ties) and an average gain of $+1.9\%$, peaking at $+3.3\%$ for Qwen3-VL-8B on MM-VisHal. Conversely, the vision-side constraint exhibits a strong bias toward anatomy-related questions, delivering the best Acc-A in $10$ of $12$ settings. Finally, the joint strategy excels in radiology-feature questions, attaining the best Acc-R in $9$ of $12$ settings. The domain-specific benefits of interventions make dynamic routing essential for superior robustness.

\begin{figure}[t]
    \centering
    \includegraphics[width=1.0\linewidth]{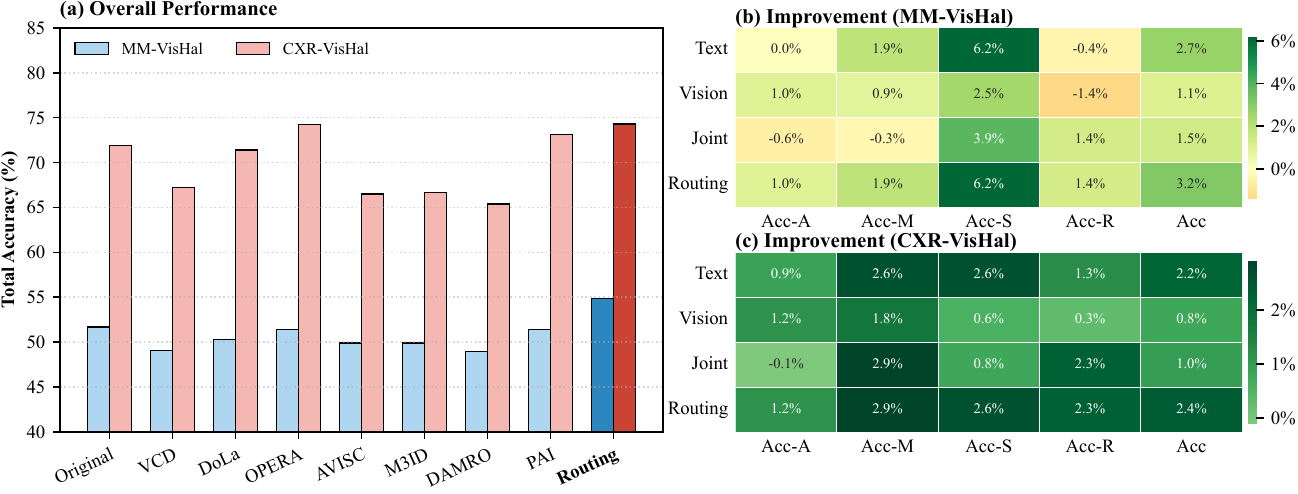}
    \caption{Evaluation of mitigation strategies. (a) Overall accuracy comparison on MM-VisHal and CXR-VisHal datasets. (b)-(c) Absolute metric improvements of individual components and our Routing over the LLaVA-Med-1.5-7B baseline.}
    \label{fig:close_mitigation_comparison}
\end{figure}

As shown in Fig.~\ref{fig:close_mitigation_comparison}, our task-aware Router consistently outperforms less robust baselines. Each module exhibits selective advantages for different questions, By dynamically exploiting these complementary strengths, routing achieves the highest overall accuracy improvements ($+3.2\%$ on MM-VisHal; $+2.4\%$ on CXR-VisHal), demonstrating superior robustness over any fixed intervention.

\begin{tcolorbox}[
  colback=gray!15,      
  colframe=black!40,    
  boxrule=0.6pt,        
  arc=6pt,              
  left=8pt,right=8pt,top=1pt,bottom=1pt, 
  before skip=2pt,      
  after skip=2pt,        
  fontupper=\small
]
\textbf{Obs 1}: Dynamic dual-side routing transcends the fragility of post-hoc corrections, consistently mitigating multi-source hallucinations.
\end{tcolorbox}

\begin{figure}[!t]
    \centering
    \includegraphics[width=1.0\linewidth]{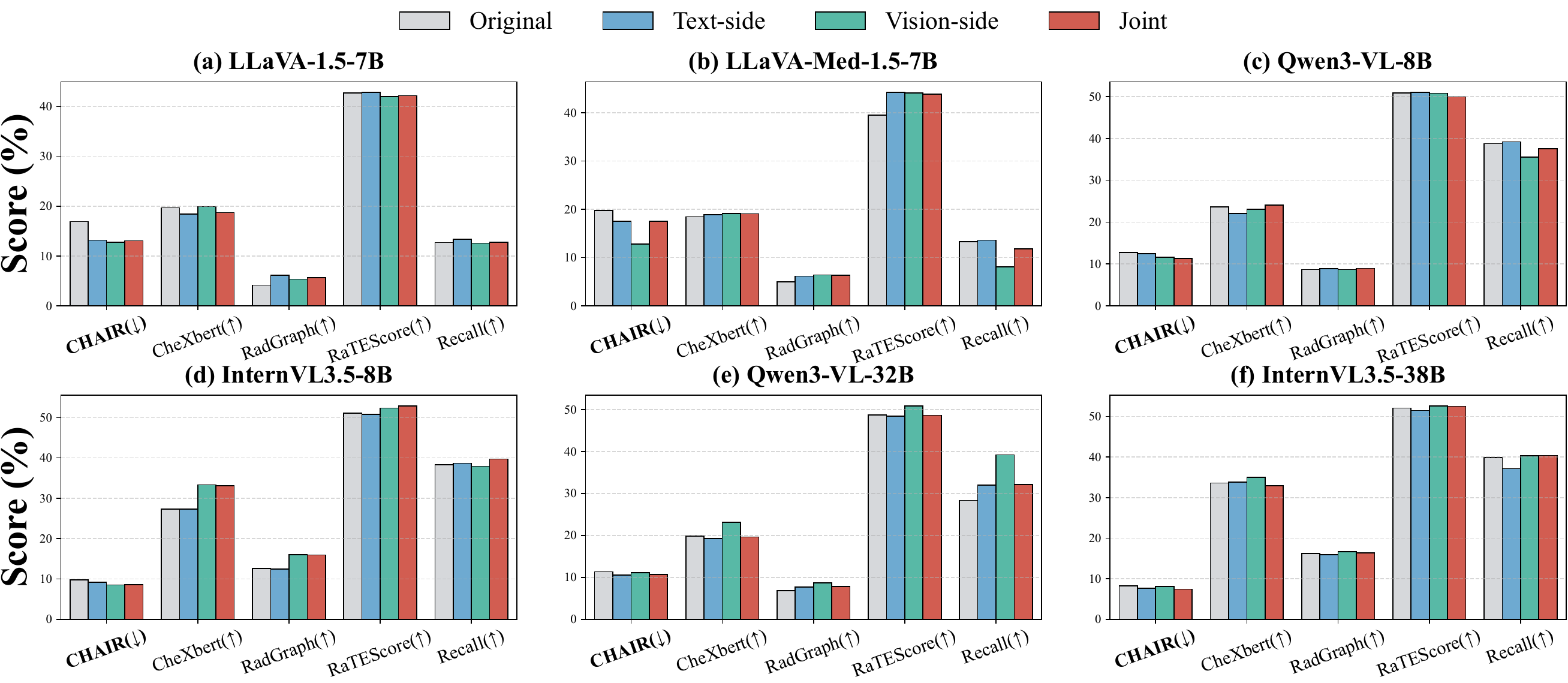}
    \caption{Open-ended evaluation results on the MIMIC-CXR for report generation. Subfigures (a) to (f) illustrate the evaluation results across six different MLLMs.}
    \label{fig:open_models_metrics_comparison}
\end{figure}

\paragraph{\textbf{Open-ended Evaluation.}}
Fig.~\ref{fig:open_models_metrics_comparison} shows that ROI-based training-free interventions reduce hallucinations with metric-dependent trade-offs: vision-side constraints most consistently lower CHAIR and improve clinical consistency, text-side injection better preserves coverage and fluency (higher RaTEScore/Recall), and the joint strategy often provides the best balance, yielding strong Recall with competitive CHAIR (notably on InternVL3.5).

\begin{figure}[t]
    \centering
    \includegraphics[width=1.0\linewidth]{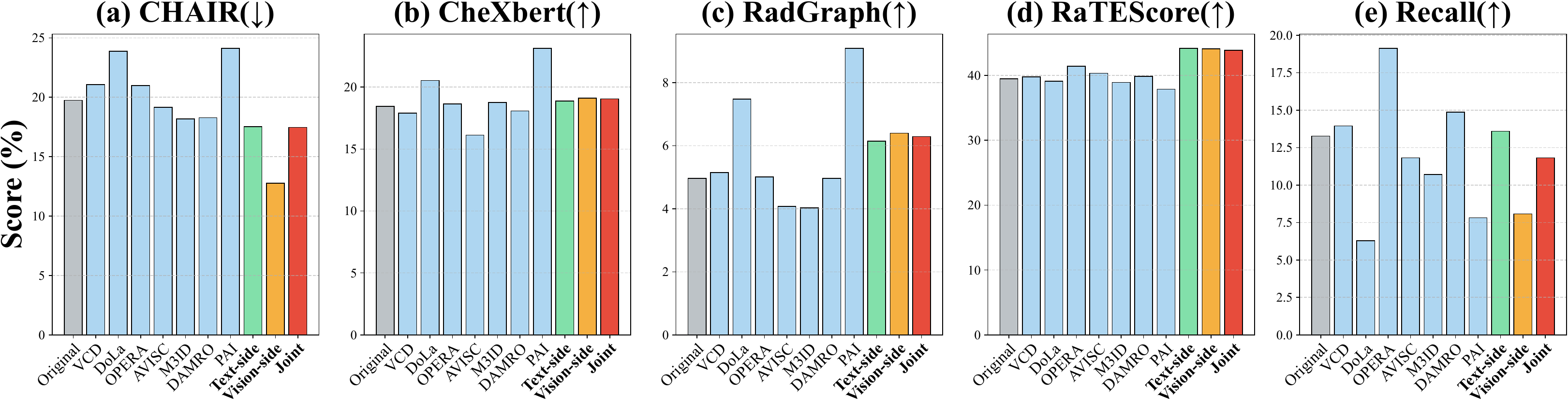}
    \caption{Open-ended mitigation comparison on MIMIC-CXR using LLaVA-Med-1.5-7B.}
    \label{fig:open_mitigation_comparison}
\end{figure}

\begin{figure}[!t]
  \centering
  \includegraphics[width=\textwidth]{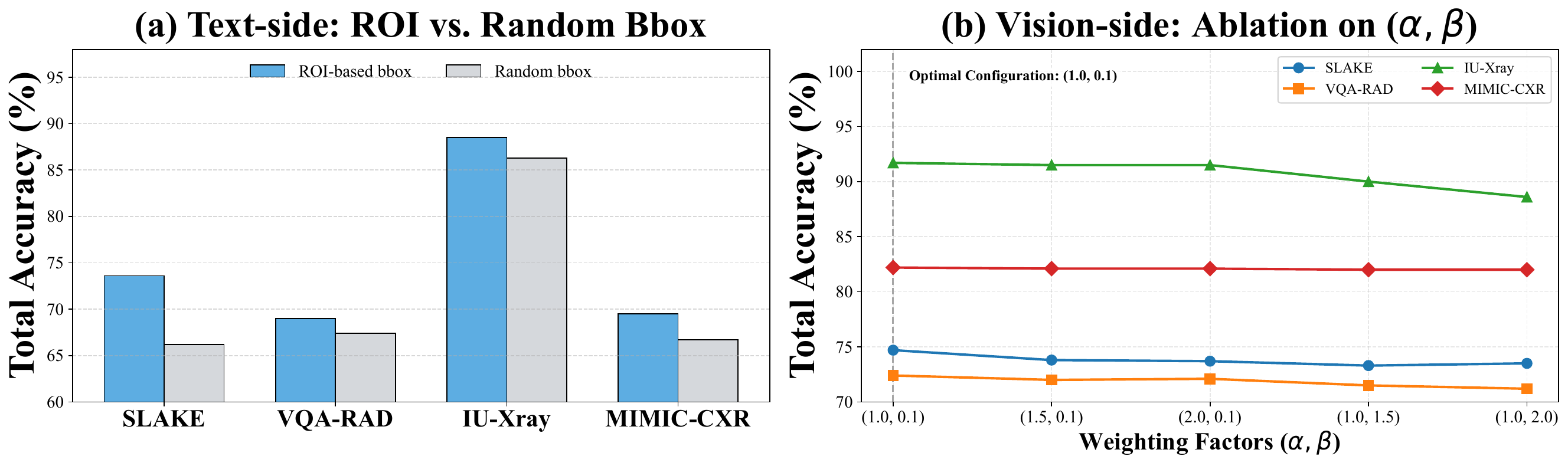}
  \caption{Controlled comparisons for text-side and vision-side methods. (a) Performance comparison between ROI-based and random bounding boxes for Qwen3-VL-8B. (b) Ablation study of weighting factors $(\alpha, \beta)$ for InternVL3.5-8B. }
  \label{fig:tv_controlled_comparison}
\end{figure}

\newlength{\ImgH}
\setlength{\ImgH}{3.25cm}

\newlength{\TxtH}
\setlength{\TxtH}{2.2cm} 

\newcommand{\ImgCell}[1]{%
  \begin{minipage}[t]{\linewidth}
    \centering
    \adjustbox{min size={\linewidth}{\ImgH},max size={\linewidth}{\ImgH},clip,center}{%
      \includegraphics{#1}%
    }
  \end{minipage}%
}

\newcommand{\TxtImgCell}[1]{%
  \begin{minipage}[t]{\linewidth}
    \begin{tcolorbox}[
      enhanced,
      colback=white,
      colframe=black!55,
      boxrule=0.5pt,
      arc=0mm,
      left=1mm,right=1mm,top=0mm,bottom=0mm,
      boxsep=0mm,
      width=\linewidth,
      height=\TxtH,      
      valign=center      
    ]
      \centering
      \includegraphics[width=\linewidth,height=\dimexpr\TxtH-2pt\relax,keepaspectratio]{#1}%
    \end{tcolorbox}
  \end{minipage}%
}

\begin{figure}[!t]
\centering
\setlength{\tabcolsep}{1.5pt}
\renewcommand{\arraystretch}{1.0}

\begin{tabular}{@{}p{0.245\linewidth}p{0.245\linewidth}p{0.245\linewidth}p{0.245\linewidth}@{}}

\multicolumn{1}{c}{\textbf{Baseline}} &
\multicolumn{1}{c}{\textbf{Text-side}} &
\multicolumn{1}{c}{\textbf{Vision-side}} &
\multicolumn{1}{c}{\textbf{Joint}}\\[-2pt]

\ImgCell{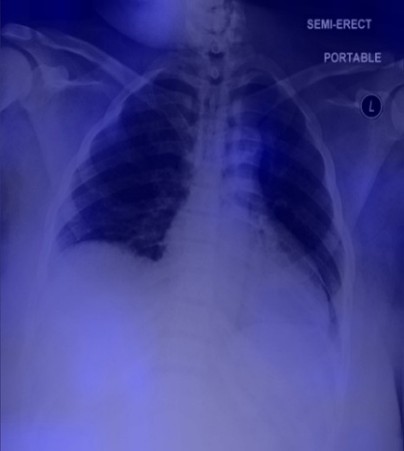} &
\ImgCell{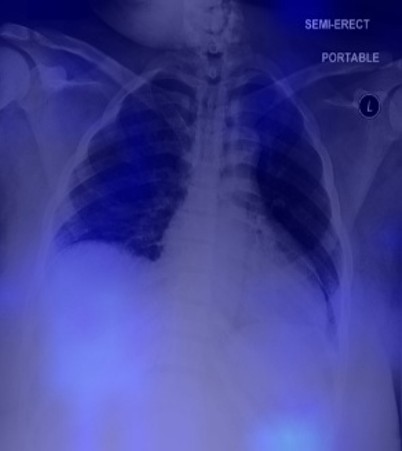} &
\ImgCell{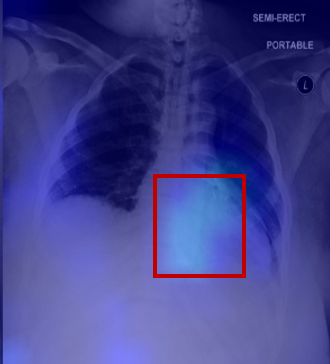} &
\ImgCell{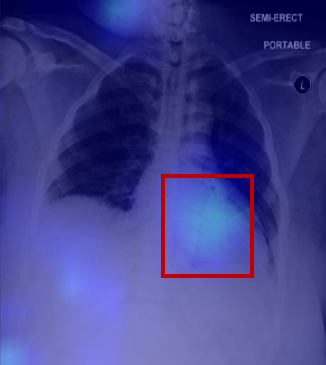} \\[-2pt]

\TxtImgCell{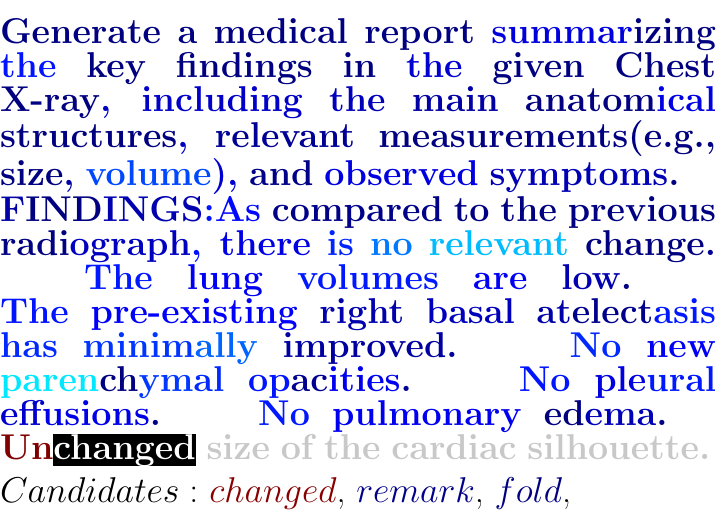} &
\TxtImgCell{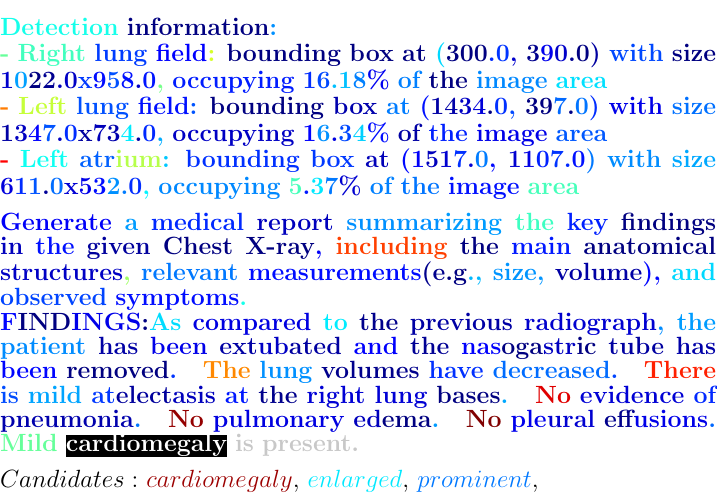} &
\TxtImgCell{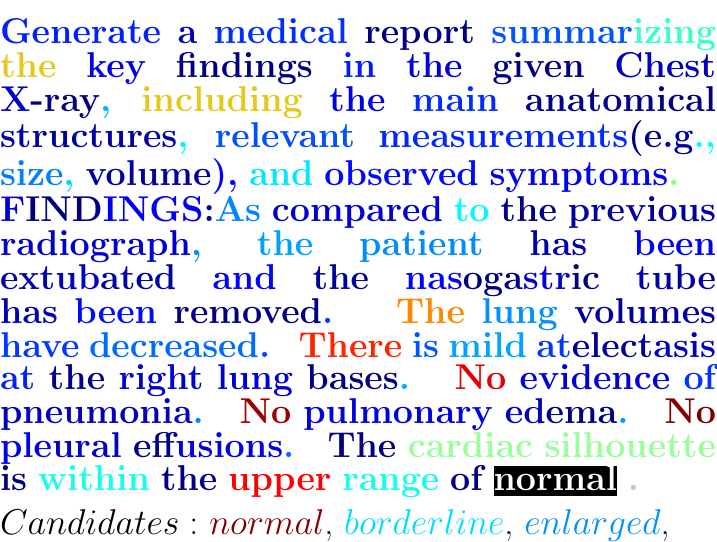} &
\TxtImgCell{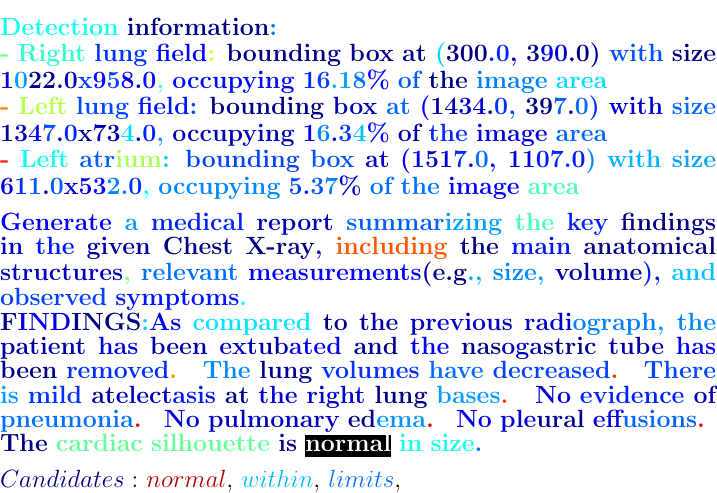} 

\end{tabular}

\caption{Token Activation Map (TAM) for visualization presents high-quality localization results on MIMIC-CXR datasets under the InternVL3.5-8B.}
\label{fig:tam}
\end{figure}

Fig.~\ref{fig:open_mitigation_comparison} highlights the trade-offs of our ROI-based interventions against baselines. While OPERA yields the highest Recall (19.13\%) and PAI excels in clinical metrics (CheXbert 23.13\%, RadGraph 9.09\%), both severely increase hallucinations. Our vision-side intervention achieves the lowest CHAIR (12.76\%) and improves clinical consistency by curbing spurious visual findings, albeit at a slight cost to Recall. Meanwhile, our text-side intervention maintains Recall and boosts RaTEScore to 44.18\%, proving structured textual injection enhances narrative quality and coherence while effectively mitigating hallucinations.

\begin{tcolorbox}[
  colback=gray!15,      
  colframe=black!40,    
  boxrule=0.6pt,        
  arc=6pt,              
  left=8pt,right=8pt,top=1pt,bottom=1pt, 
  before skip=2pt,      
  after skip=2pt,        
  fontupper=\small
]
\textbf{Obs 2}: The central challenge in open-ended evaluation remains the difficulty of balancing report completeness and hallucination suppression.

\end{tcolorbox}

\subsection{Ablation Study and Qualitative Analysis}
\paragraph{\textbf{Ablation Study.}}Fig.~\ref{fig:tv_controlled_comparison} illustrates the ablation studies for both interventions. \textbf{(1) Text-side:} Random bbox degrade performance across all benchmarks, proving that text-side improvements rely on the injected spatial grounding of ROI coordinates. \textbf{(2) Vision-side:} In vision-side weighting, $(\alpha=1.0, \beta=0.1)$ is the optimal setting. Over-amplifying ROI features ($\alpha$) provides no gain, while increasing non-ROI weights ($\beta$) consistently introduces distracting signals.

\paragraph{\textbf{Qualitative Analysis.}} Fig.~\ref{fig:tam} provide TAM visualizations to elucidate how our interventions steer the grounding mechanism. The baseline’s diffuse attention during the generation of \emph{``changed''} is sharpened by vision-side injection, which refocuses the model on the left atrial region—a key semantic anchor for the ground-truth \emph{``normal''} heart size.

\section{Conclusion}
We address visual misinterpretation hallucinations in medical MLLMs with a training-free, evidence-injection framework. Using MedSAM-extracted ROI priors as verifiable anatomical evidence, our method improves visual grounding and clinical consistency through text-side evidence injection, vision-side activation recalibration, and their joint strategy. We further introduce a lightweight decoupled router that dynamically selects modality-specific interventions based on task semantics. Extensive evaluations on multiple MLLMs across close-ended and open-ended tasks, supported by controlled ablations and visualizations, show robust hallucination mitigation over existing baselines while improving reasoning accuracy and factual reliability. Overall, this work provides a practical path toward visually grounded and controllable generation in clinical workflows.

\begin{credits}
\subsubsection{\ackname} The work was supported by the National Key R\&D Program of China under Grant 2025YFE0216500, the Major Program of National Natural Science Foundation of China under Grant 62595802, the Foundation for Outstanding Research Groups of Hubei Province of China under Grant 2025AFA012, the 111 Project on Computational Intelligence and Intelligent Control under Grant B18024, and the National Research Foundation, Prime Minister's Office, Singapore under its IN-CYPHER Campus for Research Excellence and Technological Enterprise (CREATE) Programme.

\subsubsection{\discintname}
The authors have no competing interests in the paper.
\end{credits}
%
%


\begin{thebibliography}{99}

\bibitem{chen2024chexagent}
Chen, Z., Varma, M., Delbrouck, J.-B., Paschali, M., Blankemeier, L., Van~Veen, D., et al.:
CheXagent: Towards a foundation model for chest X-ray interpretation.
In: AAAI 2024 Spring Symposium Series (Clinical Foundation Models). OpenReview.net (2024).
\url{https://openreview.net/forum?id=P3LOmrZWGR}

\bibitem{moor2023medflamingo}
Moor, M., Huang, Q., Wu, S., Yasunaga, M., Dalmia, Y., Leskovec, J., Zakka, C., Reis, E.P., Rajpurkar, P.:
Med-Flamingo: A multimodal medical few-shot learner.
In: Proceedings of the 3rd Machine Learning for Health Symposium (ML4H 2023).
Proceedings of Machine Learning Research, vol. 225, pp. 353--367. PMLR (2023)

\bibitem{thawakar2024xraygpt}
Thawakar, O.C., Shaker, A.M., Mullappilly, S.S., Cholakkal, H., Anwer, R.M., Khan, S., Laaksonen, J., Khan, F.:
XrayGPT: Chest radiographs summarization using large medical vision-language models.
In: Proceedings of the 23rd Workshop on Biomedical Natural Language Processing (BioNLP 2024),
pp. 440--448. Association for Computational Linguistics (2024)

\bibitem{wu2025radfm}
Wu, C., Zhang, X., Zhang, Y., Hui, H., Wang, Y., Xie, W.:
Towards generalist foundation model for radiology by leveraging web-scale 2D\&3D medical data.
Nature Communications \textbf{16}, 7866 (2025)

\bibitem{lee2024llmcxr}
Lee, S., Kim, W.J., Chang, J., Ye, J.C.:
LLM-CXR: Instruction-finetuned LLM for CXR image understanding and generation.
In: International Conference on Learning Representations (ICLR) (Poster) (2024).
OpenReview.net. \url{https://openreview.net/forum?id=BqHaLnans2}

\bibitem{jing2025tutorialhallucinations}
Jing, L., Zhang, Y., Du, X.:
Tutorial proposal: Hallucinations in large language models and large vision-language models.
In: Proceedings of the 2025 International Conference on Multimedia Retrieval (ICMR),
pp. 2138--2139 (2025)

\bibitem{lin2024trustworthyreview}
Lin, Z., Guan, S., Zhang, W., Zhang, H., Li, Y., Zhang, H.:
Towards trustworthy LLMs: A review on debiasing and dehallucinating in large language models.
Artificial Intelligence Review \textbf{57}(9) (2024)

\bibitem{tu2025attentionreallocation}
Tu, C., Ye, P., Zhou, D., Bai, L., Yu, G., Chen, T., Ouyang, W.:
Attention reallocation: Towards zero-cost and controllable hallucination mitigation of MLLMs.
International Journal of Computer Vision \textbf{134}(1) (2025)

\bibitem{chang2025medheval}
Chang, A., Huang, L., Bhatia, P., Kass-Hout, T., Ma, F., Xiao, C.:
MedHEval: Benchmarking hallucinations and mitigation strategies in medical large vision-language models.
arXiv:2503.02157 (2025)

\bibitem{zhu2025trust}
Zhu, Z., Zhang, Y., Zhuang, X., Zhang, F., Wan, Z., Chen, Y., Long, Q., Zheng, Y., Wu, X.:
Can we trust AI doctors? A survey of medical hallucination in large language and large vision-language models.
In: Findings of the Association for Computational Linguistics: ACL 2025,
pp. 6748--6769. Association for Computational Linguistics (2025)

\bibitem{asgari2025clinicalsafety}
Asgari, E., Monta\~{n}a-Brown, N., Dubois, M., et al.:
A framework to assess clinical safety and hallucination rates of LLMs for medical text summarisation.
npj Digital Medicine \textbf{8}, 274 (2025)

\bibitem{leng2024vcd}
Leng, S., Zhang, H., Chen, G., Li, X., Lu, S., Miao, C., Bing, L.:
Mitigating object hallucinations in large vision-language models through visual contrastive decoding.
In: Proceedings of the IEEE/CVF Conference on Computer Vision and Pattern Recognition (CVPR),
pp. 13872--13882 (2024)

\bibitem{chuang2023dola}
Chuang, Y.-S., Xie, Y., Luo, H., Kim, Y., Glass, J., He, P.:
DoLa: Decoding by contrasting layers improves factuality in large language models.
arXiv:2309.03883 (2023)

\bibitem{huang2024opera}
Huang, Q., Dong, X., Zhang, P., Wang, B., He, C., Wang, J., Lin, D., Zhang, W., Yu, N.:
OPERA: Alleviating hallucination in multi-modal large language models via over-trust penalty and retrospection-allocation.
In: Proceedings of the IEEE/CVF Conference on Computer Vision and Pattern Recognition (CVPR),
pp. 13418--13427 (2024)

\bibitem{woo2024avisc}
Woo, S., Kim, D., Jang, J., Choi, Y., Kim, C.:
Don't miss the forest for the trees: Attentional vision calibration for large vision language models.
arXiv:2405.17820 (2024)

\bibitem{gong2024damro}
Gong, X., Ming, T., Wang, X., Wei, Z.:
DAMRO: Dive into the attention mechanism of LVLM to reduce object hallucination.
In: Proceedings of the 2024 Conference on Empirical Methods in Natural Language Processing (EMNLP),
pp. 7696--7712 (2024)

\bibitem{liu2025pai}
Liu, S., Zheng, K., Chen, W.:
Paying more attention to image: A training-free method for alleviating hallucination in LVLMs.
In: European Conference on Computer Vision (ECCV), pp. 125--140. Springer (2025)

\bibitem{favero2024m3id}
Favero, A., Zancato, L., Trager, M., Choudhary, S., Perera, P., Achille, A., Swaminathan, A., Soatto, S.:
Multi-modal hallucination control by visual information grounding.
In: Proceedings of the IEEE/CVF Conference on Computer Vision and Pattern Recognition (CVPR),
pp. 14303--14312 (2024)

\bibitem{liu2021slake}
Liu, B., Zhan, L.-M., Xu, L., Ma, L., Yang, Y., Wu, X.-M.:
SLAKE: A semantically-labeled knowledge-enhanced dataset for medical visual question answering.
In: 2021 IEEE 18th International Symposium on Biomedical Imaging (ISBI). IEEE,
pp. 1650--1654 (2021)

\bibitem{lau2018vqarad}
Lau, J.J., Gayen, S., Ben Abacha, A., Demner-Fushman, D.:
A dataset of clinically generated visual questions and answers about radiology images.
Scientific Data \textbf{5}(1), 1--10 (2018)

\bibitem{demner2016iu}
Demner-Fushman, D., Kohli, M.D., Rosenman, M.B., Shooshan, S.E., Rodriguez, L.,
Antani, S., Thoma, G.R., McDonald, C.J.:
Preparing a collection of radiology examinations for distribution and retrieval.
Journal of the American Medical Informatics Association \textbf{23}(2), 304--310 (2016)

\bibitem{johnson2019mimiccxr}
Johnson, A.E.W., Pollard, T.J., Berkowitz, S.J., Greenbaum, N.R., Lungren, M.P.,
Deng, C.-Y., Mark, R.G., Horng, S.:
MIMIC-CXR, a de-identified publicly available database of chest radiographs with free-text reports.
Scientific Data \textbf{6}(1), 317 (2019)

\bibitem{liu2024llava15}
Liu, H., Li, C., Li, Y., Lee, Y.J.:
Improved baselines with visual instruction tuning.
In: Proceedings of the IEEE/CVF Conference on Computer Vision and Pattern Recognition (CVPR),
pp. 26296--26306 (2024)

\bibitem{li2023llavamed}
Li, C., Wong, C., Zhang, S., Usuyama, N., Liu, H., Yang, J.-W., Naumann, T., Poon, H., Gao, J.:
LLaVA-Med: Training a large language-and-vision assistant for biomedicine in one day.
In: Advances in Neural Information Processing Systems, vol. 36, pp. 28541--28564.
Curran Associates, Inc. (2023)

\bibitem{bai2025qwen3vl}
Bai, S., Cai, Y., Chen, R., Chen, K., Chen, X., Cheng, Z., Deng, L., Ding, W., Gao, C., Ge, C., et al.:
Qwen3-VL technical report.
arXiv preprint arXiv:2511.21631 (2025)

\bibitem{wang2025internvl35}
Wang, W., Gao, Z., Gu, L., Pu, H., Cui, L., Wei, X., Liu, Z., Jing, L., Ye, S., Shao, J., et al.:
InternVL3.5: Advancing open-source multimodal models in versatility, reasoning, and efficiency.
arXiv preprint arXiv:2508.18265 (2025)

\end{thebibliography}
\end{document}